\documentclass[letterpaper, 10 pt, conference]{ieeeconf}

\IEEEoverridecommandlockouts                              

\usepackage{xcolor}
\usepackage{graphics} 
\usepackage{graphicx}
\usepackage{epsfig} 
\usepackage{subfigure}
\newtheorem{remark}{Remark}

\usepackage{hyperref}
\usepackage{amsmath} 
\usepackage{amssymb}  
\usepackage[ruled,vlined]{algorithm2e}
\usepackage{cite}

\title{\LARGE \bf
Learning Dynamic Manipulation Skills from Haptic-Play
}

\author{Taeyoon Lee$^{1}$, Donghyun Sung$^{2}$, Kyoungyeon Choi$^{2}$, Choongin Lee$^{1}$, Changwoo Park$^{1}$ and Keunjun Choi$^{1}$
\thanks{$^{1}$Taeyoon Lee, Choongin Lee, Changwoo Park and Keunjun Choi are with NAVER LABS, Seongnam, Korea. 
        {\tt\small ty-lee@naverlabs.com}, {\tt\small l.ci@naverlabs.com}, {\tt\small changwoo.park@naverlabs.com}, {\tt\small keunjun.choi@naverlabs.com}
        }%
\thanks{$^{2}$Donghyun Sung and Kyoungyeon Choi have conducted this research during a robotics internship program at NAVER LABS. {\tt\small sdh1259@snu.ac.kr}, {\tt\small kenbaxter1221@gmail.com}
}
}

\begin{document}

\maketitle
\thispagestyle{empty}
\pagestyle{empty}

\begin{abstract}
In this paper, we propose a data-driven skill learning approach to solve highly dynamic manipulation tasks entirely from offline teleoperated play data. We use a bilateral teleoperation system to continuously collect a large set of dexterous and agile manipulation behaviors, which is enabled by providing direct force feedback to the operator.
We jointly learn the state conditional latent skill distribution and skill decoder network in the form of goal-conditioned policy and skill conditional state transition dynamics using a two-stage generative modeling framework. This allows one to perform robust model-based planning, both online and offline planning methods, in the learned skill-space to accomplish any given downstream tasks at test time. We provide both simulated and real-world dual-arm box manipulation experiments showing that a sequence of force-
controlled dynamic manipulation skills can be composed in
real-time to successfully configure the box to the randomly
selected target position and orientation;  please refer to the supplementary video \url{https://youtu.be/LA5B236ILzM}.

\end{abstract}


\section{INTRODUCTION}
An intelligent robot performing manipulation tasks in complex, unstructured environments should be provided with sufficiently sophisticated and diverse set of programmable {\em skills} beyond motion; as M. Mason notes in \cite[p. 4]{mason2018toward}:
\begin{quote}
    Manipulation refers to an agent’s control of its environment through selective contact. ... interpret “control” broadly enough to include motion, as well as processes such as wiping, polishing, painting, and burnishing ... .
\end{quote}

Manipulation problem poses a unique, important set of challenges in decision making and control. Not to mention all the difficulties in sequential reasoning and planning problems complicated in large part to the hybrid, complex, mode-switching nature of the underlying dynamics \cite{ruggiero2018nonprehensile}, a general task description or labeling for each of the diverging spectrum of motor behaviors associated with the task is more often than not less straightforward. For these and other reasons, robot manipulation remains a practical and theoretical challenge.

In recent years, data-driven learning approaches have shown great promise toward achieving the ultimate goal of building a general-purpose robot. Not only have robots shown to master very complex tasks through repeated trial-and-error \cite{akkaya2019solving, hwangbo2019learning}, but more importantly, they are beginning to solve a more broad spectrum of tasks by making use of a repertoire of general-purpose skills learned from a database of prior experiences. 

Of course, a learned set of skills should be only as good as the richness and diversity of data. In general, collecting a sufficient amount of high-quality data samples in real robotic hardware raises many practical concerns, including safety issues.  Some notable recent works  \cite{lynch2020learning, gupta2019relay, mandlekar2018roboturk} have shown that an extensive collection of offline manipulation datasets can be gathered from (crowd-sourced) human demonstrations using a VR teleoperation system. 
Meanwhile, it is to note that one common key to practical success in large-scale skill learning is intimately related to how one defines skills in the first place; they should be defined in such a way that the required training dataset can be collected and labeled at scale. For instance, one popular choice of skill definitions is goal-conditioned action trajectories or policies which can be self-supervised at scale by relabeling the goals as the final states within sliding windows of trajectory database \cite{lynch2020learning, gupta2019relay, ajay2020opal,chebotar2021actionable}.
There can be many forms of which skills are composed to solve the given task. For instance, skills can be explicitly composed by selecting the subsequent skills over time; the planning strategy can be directly performed online \cite{fang2019dynamics,sharma2019dynamics,ichter2020broadlyexploring}, learned through repeated online interactions \cite{gupta2019relay, pertsch2020accelerating} or learned offline \cite{ajay2020opal}. Further, in \cite{peng2019mcp}, multiple skills are simultaneously composed using a multiplicative policy structure for controlling complex, high-dimensional systems, like humanoids.

\begin{figure}
\centering
\includegraphics[width=0.8\columnwidth]{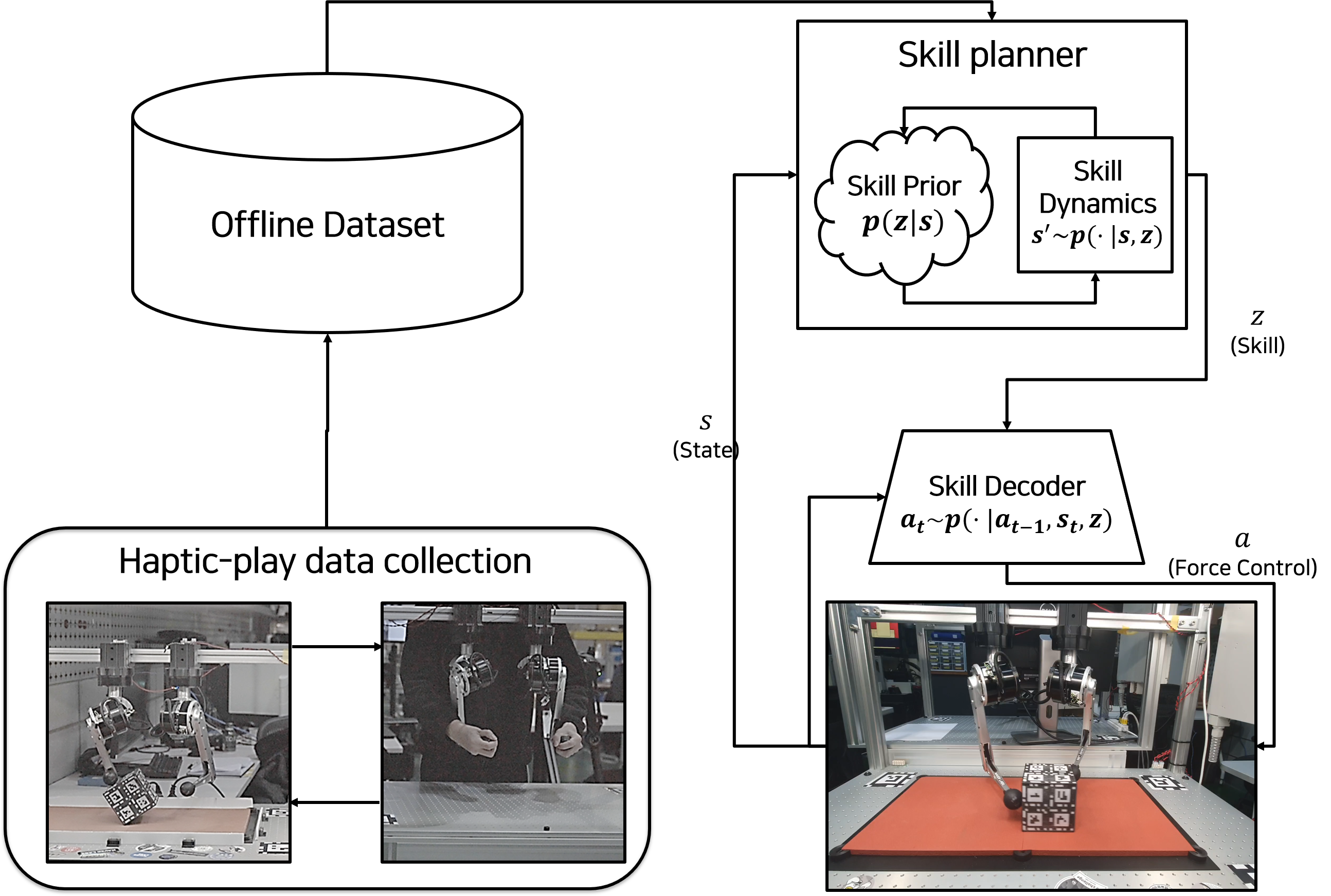}
\caption{Overall flow of the proposed offline skill learning framework.}
\label{skill_hierarhcy}
\end{figure}

In this paper, we propose an offline skill learning framework in which to solve highly dynamic and contact-rich real-world robotic manipulation tasks. We first make use of the bilateral teleoperation system to generate an extensive collection of unlabeled, agile and dexterous force-controlled manipulation behaviors. Using this offline trajectory database, we learn state-conditional latent skill distribution which jointly encodes finite-horizon goal-conditioned action trajectories and skill-conditional state transition using a two-stage conditional generative modeling framework. Then, this finally allows one to perform model-based planning, e.g., model predictive control (MPC), model-based offline RL, etc., to solve the given downstream task at test time. Our method and claims are validated via both simulated and real-world box manipulation experiments involving two sets (master and slave devices) of 8-dof dual-arm torque-controlled robot manipulators; the results demonstrate that a sequence of force-controlled dynamic manipulation skills can be composed in real-time to successfully configure the box to the randomly selected target position and orientation.

\section{PROBLEM FORMULATION}

Our main goal is to extract a reusable, generative skills as well as skill-conditioned state transition model from a large, unlabeled play dataset. At test time, these twin components are used to perform model-based planning for solving any given downstream tasks. 

To begin, let us define infinite horizon, discrete time Markov Decision Process (MDP), $\mathcal{M} = (\mathcal{S}, \mathcal{A}, p, R, \gamma,\rho_0)$, where $\mathcal{S}$ and $\mathcal{A}$ are the continuous spaces of states and actions respectively,  $p(\cdot|s,a)$ for $s \in \mathcal{S}, a\in\mathcal{A}$ is the state transition probability density, $R: \mathcal{S}\times\mathcal{A}\rightarrow\mathbb{R}$ is the reward function, $\gamma\in [0,1)$ is the discount factor, and $\rho_0(\cdot)$ is the initial state density.

Given an offline trajectory database, $\mathcal{D}\triangleq\{(s_{\mathrm{i}}, a_i)\}_{i=1}^{N}$, containing continuous and diverse motor behaviors, we aim at learning the joint distribution,
\begin{equation*}
    p(\tau_{t,W}, s_{t+W} | s_t),
\end{equation*}

of the state-action trajectory, $\tau_{t,W} \triangleq \{(s_{t+k}, a_{t+k})\}_{k=0}^{W-1}$ of some fixed window size $W$ and the goal state $s_{t+W}$ conditioned on the initial state $s_t$. For this purpose, we rearrange the original dataset $\mathcal{D}$ over the sliding windows of the entire trajectory as,  
\begin{equation*}
\mathcal{D}_{\tau,W} \triangleq \{s_t, a_t, \cdots, s_{t+W-1}, a_{t+W-1}, s_{t+W}\}_{t=1}^{N-W}\
\end{equation*}
For notation simplicity, we relabel the samples drawn from the dataset $\mathcal{D}_{\tau, W}$ as,
\begin{equation}
s_{\mathrm{i}} \leftarrow s_t, ~ s_{\mathrm{g}} \leftarrow s_{t+W} ~ \text{and}~ \tau \leftarrow \tau_{t,W},    
\end{equation}
denoting the (intermediate) initial and goal state and the state-action trajectory respectively.

We now assume a following latent variable model over the conditional distribution described above:
\begin{align}
    p(\tau, s_{\mathrm{g}}|s_{\mathrm{i}}) = \int_{\mathcal{Z}}&p(\tau|s_{\mathrm{g}},s_{\mathrm{i}},z)~p(s_{\mathrm{g}}|s_{\mathrm{i}},z)~ p(z|s_{\mathrm{i}})~dz \nonumber \\
    =\int_{\mathcal{Z}}& \prod_{(s,a,s')\in\tau}\Big\{\underbrace{\pi(a|s,s_{\mathrm{g}},z)}_{\mathrm{skill-policy}} ~ p(s'|s,a)\Big\} \nonumber\\
& ~\underbrace{p(s_{\mathrm{g}}|s_{\mathrm{i}}, z)}_{\mathrm{skill-dynamics}}~\underbrace{p(z|s_{\mathrm{i}})}_{\mathrm{skill-prior}}~dz, \label{skill_definition}
\end{align}
where any instance drawn from this distribution would represent a single realization of a {\em skill}. 

The latent variable $z\in\mathcal{Z}$ drawn from the {\bf skill-prior} distribution $p(z|s_{\mathrm{i}})$ would represent the admissible skills at the initial state $s_{\mathrm{i}}$, while the feedback {\bf skill-policy}, $\pi(a|s,s_{\mathrm{g}},z)$ for all $(s,a)\in\tau$, decodes the action sequences that achieve the goal state $s_{\mathrm{g}}$ using a particular latent skill $z$. Lastly, the {\bf skill-dynamics} would enable the robot to be aware of the goal state $s_{\mathrm{g}}$ in advance before actually executing the skill $z$ at state $s_{\mathrm{i}}$. 

Finally, provided that the skill-policy and skill-dynamics can accurately reconstruct the states and actions of the original MDP $\mathcal{M}$, we define a new skill-level MDP $\mathcal{M}_{\mathcal{Z}}= (\mathcal{S},\mathcal{Z},p_z,R_z,\gamma, \rho_0)$, for the purpose of model-based planning in the skill-space $\mathcal{Z}$. The transition probability $p_z$ stands for the skill-dynamics $p(s_{t+W}|s_t,z)$, and $R_z: \mathcal{S} \times \mathcal{Z} \rightarrow \mathbb{R}$ is the reward function defined over the state $s$ and the latent skill $z$. Our final objective is then to find the optimal skill-level policy $\pi^*(z|s)$ maximizing the cumulative reward, i.e.,
\begin{equation}
 \max_{\pi^*, s_0 \sim\rho(\cdot)} \mathbb{E}\left[\sum_{k=0}^{\infty}\gamma^k R(s_{k\cdot W}, z_k ) ~\bigg|~ s_0\right] \label{skill_MDP_prob}
\end{equation}
for solving any given downstream task.
\begin{remark} \label{rem:skill_extension}
It is noted that owing to the recurrent model-based nature of the skill definition, one can temporarily extend the duration of a unit skill while still allowing the model-based future state/trajectory prediction viable, i.e., at state $s_0$, sample $z_0 \sim p(\cdot|s_0)$ and predict $s_1 \sim p(\cdot|s_0, z_0)$, sample $z_1 \sim p(\cdot|s_1)$ and predict $s_2 \sim p(\cdot|s_1, z_1)$, and so on.
\end{remark}

\section{METHOD}
In this Section, we describe in detail how the components stated in \eqref{skill_definition} are jointly trained using the offline dataset $\mathcal{D}_{\tau, W}$, while also discussing how different generative model learning approaches can affect the effectiveness of the learned skill representations in model-based planning. We start by introducing how the offline dataset containing diverse manipulation behaviors is collected in this study.
\subsection{Haptic-Play: Offline Dataset Collection}
As depicted in Figure \ref{skill_hierarhcy}, we use a bilateral teleoperation system to collect an extensive collection of force-controlled manipulation behaviors for offline skill learning. Two sets of proprioceptive torque-controlled robots \cite{katz2019mini}, master and slave, are synchronized using low latency, high stiffness impedance control \cite{lee2006passive} so that the operator can not only transmit motion commands but also regulate force through the master device owing to the force/haptic feedback.

We argue that providing the demonstrator with the haptic feedback information greatly extends the degree to which dynamic, interactive behaviors can be demonstrated; the operator can more directly demonstrate both force and position commands that constitute natural, agile and dexterous contact-rich manipulation behaviors to the robot (slave). Also, it is to note that no manual labeling is needed in the data collection stage; that is, the operator is not repeatedly asked to generate a set of episodic demonstrations for a specific task. The operator continuously demonstrates diverse motor behaviors by simply ``playing'' \cite{lynch2020learning} the robot by means of the haptic device.

\subsection{Skill Learning Problem Formulation}
Given the offline trajectory dataset, our aim is to learn a state-conditional distribution $p(z|s_{\mathrm{i}})$ of skills that jointly realize feedback policy, $\pi(a|s, s_{\mathrm{g}}, z)$, and associated outcome state, i.e., $s_{\mathrm{g}}\sim p(s_{\mathrm{g}}|s_{\mathrm{i}},z)$.
We formulate this conditional generative skill learning problem as follows:
\begin{align}
    \min_{\theta_\pi, \theta_f, \phi, \psi}& \mathbb{E}_{ \mathcal{D}_{\tau,W}}\left[\mathbb{E}_{q_{\phi}(z|\tau, s_{\mathrm{g}}, s_{\mathrm{i}})}\left[\mathcal{L}_{\mathrm{rec}}\right]\right] \label{ours_reconstruction}\\
    \mathrm{s.t.}&~ \hat{p}_{\mathcal{D},\phi}(\cdot|s_{\mathrm{i}}) = p_{\psi}(\cdot|s_{\mathrm{i}}), ~ \mathrm{for~all}~ s_{\mathrm{i}} \in \mathcal{D}_{\tau,W} \label{ours_prior_matching} 
\end{align}
where $q_{\phi}(z|\tau, s_{\mathrm{g}}, s_{\mathrm{i}})$ is the {\bf skill posterior}, and 
\begin{equation}
\mathcal{L}_{\mathrm{rec}} \triangleq \sum_{(s,a)\in \tau}\|a - \pi_{\theta_{\pi}}(s,s_{\mathrm{g}},z)\|^2 + \|s_{\mathrm{g}} - f_{\theta_f}(s_{\mathrm{i}},z)\|^2
\end{equation}
is the skill reconstruction loss with $\pi_{\theta_{\pi}}$ and $f_{\theta_f}$ being the deterministic function of the feedback policy and skill-dynamics respectively. Further,
\begin{equation}
    \hat{p}_{\mathcal{D},\phi}(z|s_{\mathrm{i}})\triangleq \mathbb{E}_{\tau, s_{\mathrm{g}} \sim \mathcal{D}_{\tau,W}}\left[q_{\phi}(z|\tau,s_{\mathrm{g}}, s_{\mathrm{i}})\right] \label{aggregated_posterior}
\end{equation}
defines the state-conditional {\bf aggregated skill posterior} distribution \cite{makhzani2015adversarial}. The formulation closely resembles with the one of Wasserstein Autoencoder (WAE) \cite[Theorem 1]{tolstikhin2017wasserstein} \footnote{Other related approaches include  Adversarial Autoencoder (AAE) \cite{makhzani2015adversarial} and InfoVAE \cite{zhao2019infovae}.}, where the objective is also to minimize the reconstruction loss while constraining the aggregated posterior to match with some fixed prior. The important distinction we make here is that the conditional prior $p_{\psi}(z|s_{\mathrm{i}})$ is jointly learned rather than being fixed. This allows to selectively generate admissible set of skills depending on the state the robot encounters.

Meanwhile, one strong baseline for conditional generative modeling is conditional Variational Autoencoder (cVAE) \cite{sohn2015cvae, lynch2020learning}. Rather than directly enforcing the prior to match with the aggregated posterior, it regularizes each of the posteriors by minimizing the Kullback-Leibler (KL) divergence, 
\begin{equation}
\mathcal{L}_{\mathrm{KL}} = D_{\mathrm{KL}}(q_{\phi}(\cdot|\tau, s_{\mathrm{g}},s_{\mathrm{i}})\| p_{\psi}(\cdot|s_{\mathrm{i}})),
\end{equation}
between the posterior and the prior; it minimizes the following unconstrained objective function,
\begin{equation}
\mathcal{J}_{\mathrm{cVAE}} = \mathbb{E}_{ \mathcal{D}_{\tau,W}}\left[\mathbb{E}_{q_{\phi}(z|\tau, s_{\mathrm{g}}, s_{\mathrm{i}})}\left[\mathcal{L}_{\mathrm{rec}}\right] + \beta\mathcal{L}_{KL}\right], \label{cvae_loss}
\end{equation}
where $\beta$ is a non-negative regularization factor balancing the reconstruction loss in relative to the prior matching loss \cite{higgins2016beta}. If $\beta \gg 1$, the posteriors are likely to collapse to the unconditional prior, leading to poor reconstruction. In contrast, if $\beta$ is set too small, it will only try to reconstruct the skill samples drawn from the aggregated posterior distribution \eqref{aggregated_posterior}, while a large portion of the skills sampled from the prior $p_\psi(\cdot|s_{\mathrm{i}})$ will be the out-of-distribution ones that are subject to poor reconstruction quality. Hence, setting the appropriate value of $\beta$ is crucial in practice. Also, VAE typically suffers from limited modeling choice confined to conditional Gaussians.

The WAE-style formulation \eqref{ours_reconstruction},\eqref{ours_prior_matching} alleviates many of the aforementioned difficulties in VAE by directly minimizing the distance $D(\cdot, \cdot)$ between the prior and aggregated skill posterior together with the reconstruction loss, i.e.,
\begin{align}
\mathcal{J}_{\mathrm{cWAE}} = &\mathbb{E}_{ \mathcal{D}_{\tau,W}}\left[\mathbb{E}_{q_{\phi}(z|\tau, s_{\mathrm{g}}, s_{\mathrm{i}})}\left[\mathcal{L}_{\mathrm{rec}}\right]\right] \nonumber\\
& + \lambda\mathbb{E}_{s_{\mathrm{i}} \sim \mathcal{D}_{\tau,W}}\left[D(\hat{p}_{\mathcal{D},\phi}(\cdot|s_{\mathrm{i}}), p_{\psi}(\cdot|s_{\mathrm{i}}))\right]
, \label{cwae_loss}
\end{align}
with large penalty factor $\lambda$\footnote{observe that the original distribution matching constraint \eqref{ours_prior_matching} is appended as the distance penalty for unconstrained optimization.}. Since the direct evaluation of the distance $D$ is intractable, the existing approaches \cite{makhzani2015adversarial,zhao2019infovae} typically rely on sampled distance estimates using, e.g., Generative Adversarial Network (GAN) or Maximum Mean Discrepancy (MMD) loss.
However, we find the use of existing approaches less straightforward for our ``conditional'' generative learning problem. Adversarial learning based on GAN can become very unstable due in large to the non-stationary nature of the training process; here in our case, the distributions of both the source and target samples, indicating the aggregated posterior and prior, are dynamically updated during the training process\footnote{fixed prior is typically used in GAN-WAE and AAE}. While, in contrast, MMD provides direct distance estimate between two sample sets, its quadratic computational cost limits the efficient use with large offline dataset. 

\subsection{Two-Stage Skill Learning Algorithm}
Inspired by the work of \cite{dai2019diagnosing}, we propose a two-stage method to solve the constrained optimization problem stated in \eqref{ours_reconstruction}, \eqref{ours_prior_matching}. The core idea is to decompose the complex skill prior matching condition \eqref{ours_prior_matching} from the reconstruction minimization \eqref{ours_reconstruction} so that each of the optimization problem becomes much more tractable with its solution stably converging to the optimum.

\subsubsection{Dimension Reduction with cVAE}
We first minimize the cVAE objective \eqref{cvae_loss} with sufficiently small selection of $\beta$ (see Algorithm \ref{alg:two_stage}). This is to make sure that the low-dimensional embedding of skill samples $z$ drawn from the aggregrated posterior are subject to sufficiently accurate reconstruction of the sequences of actions and the goal state. 
The main reason why we don't use vanilla Autoencoder that comes with deterministic latent embedding and rather regularize the posterior distribution with minimal KL loss ($\beta \sim 0.001$) is to prevent the target data distribution, $\hat{p}_{\mathcal{D},\phi}$, from being overly complex or having degenerate support, and stabilize the maximum likehood training conducted in the second stage.

\begin{figure}
    \centerline{
    \subfigure[CVAE ($\beta \ll 1$)]
    {
    \includegraphics[width=0.3\columnwidth]{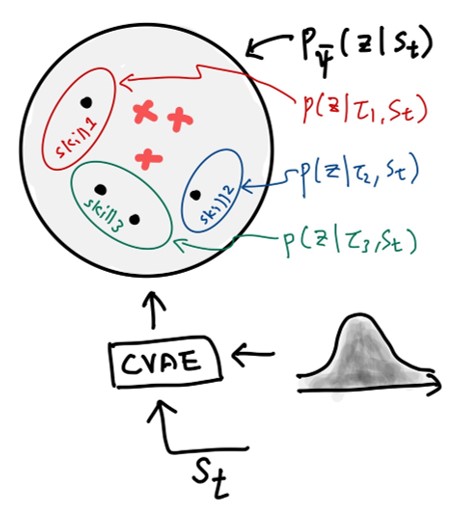}%
    \label{fig:cvae}}
    \hfil
    \subfigure[Flow]
    {
    \includegraphics[width=0.3\columnwidth]{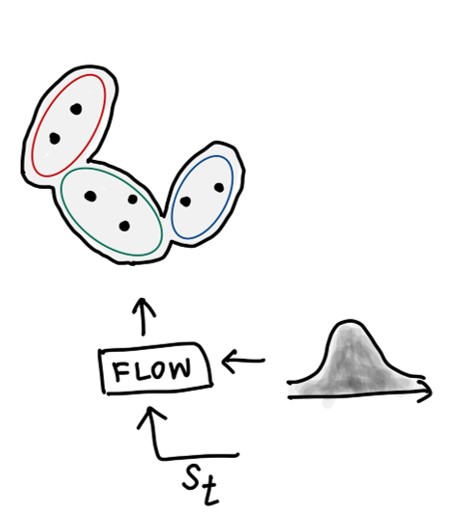}%
    \label{fig:flow}}}
\caption{Pictorial description of the improvement in the state-conditional prior matching performance from CVAE (first stage) to flow-based generative learning (second stage). The black bold dots indicate the in-distribution skill samples drawn from the prior $p_{\psi}(\cdot|s_t)$, while the red crosses indicate the out-of-distribution ones.}
\label{fig:cvae_vs_flow}
\end{figure}
\subsubsection{Generative Skill Learning}
Our next goal is to replace the simple conditional Gaussian skill prior model from the first stage cVAE training with a sufficiently complex, flexible skill prior model $p_{\psi}(\cdot|s_{\mathrm{i}})$. More specifically, we fit the the generative skill distribution $p_{\psi}(\cdot|s_{\mathrm{i}})$ to the aggregated distribution $\hat{p}_{\mathcal{D},\phi}(\cdot|s_{\mathrm{i}})$ of the fixed skill posteriors $q_{\phi}(\cdot|\tau,s_{\mathrm{g}}, s_{\mathrm{i}})$ pretrained in the first stage via maximum likelihood learning:
\begin{equation}
    \min_{\psi} \mathcal{J}_{\mathrm{ML}}= \mathbb{E}_{\mathcal{D}_{\tau,W}}\left[\mathbb{E}_{q_{\phi}(z| \tau, s_{\mathrm{g}},s_{\mathrm{i}})}\left[ -\log p_{\psi}(z|s_{\mathrm{i}})\right]\right]. \label{flow_ml_loss}
\end{equation}
This is equivalent to KL divergence minimization, i.e.,
\begin{align}
\mathcal{J}_{\mathrm{ML}} =& \mathbb{E}_{s_\mathrm{i}\sim\mathcal{D}_{\tau,W}}\left[D_{\mathrm{KL}}(\hat{p}_{\mathcal{D},\phi}(\cdot|s_{\mathrm{i}}) \| p_{\psi}(z|s_{\mathrm{i}}))\right] \nonumber\\ &\quad+ \underbrace{\mathbb{E}_{s_\mathrm{i}\sim\mathcal{D}_{\tau,W}}[\mathcal{H}(\hat{p}_{\mathcal{D},\phi}(\cdot|s_{\mathrm{i}}))]}_{\text{constant conditional entropy}}.
\end{align}

We choose to use flow-based models for this purpose, as it is an unique class of deep generative model that allows both direct density evaluation and fast forward sampling through simple forward passes \cite{kingma2016improved}. In flow-based models, one assumes a sample $z$ is generated via direct diffeomorphic transformation $h_{\psi}$ of the base sample $u \sim p_0(\cdot)$, i.e.,$z = h_{\psi}(u; s)$
, where the base distribution $p_0$ is typically set as the standard normal or uniform distribution. Density evaluation can be simply given as $p(z|s) = p_0(u)\left |\det\left(\frac{\partial }{\partial z}h_{\psi}^{-1}(z; s)\right)\right|$. The training becomes very stable owing to the direct density evaluation in maximum likelihood learning. This results in the second stage skill learning described in Algorithm \ref{alg:two_stage}.\

We argue that direct evaluation of the likelihood loss enabled by flow-based models is a major practical appeal over other models such as GAN, as one can consistently estimate and compare the distribution matching quality among different instances of models based on the likelihood loss. As depicted in Figure \ref{fig:cvae_vs_flow}, this allows one to explicitly control the amount of distribution match or out-of-distribution sampling when generating skills from the learned conditional prior model. We would verify later in the experimental Section how this loss metric strongly correlates with the downstream task performance.
Lastly, as will be discussed in the following subsection, we highlight that for skill sample generation to be conducted in the context of real-time, model-based planning, fast and parallel sampling is highly desirable. In this paper, we use Inverse Autoregessive Flow (IAF) network \cite{kingma2016improved} for enabling fast forward sampling.

\begin{algorithm}[t]
\DontPrintSemicolon
\SetAlgoLined
\SetKwInOut{Input}{Input}
\SetKwInOut{Output}{Output}
\Input{Offline trajectory dataset $\mathcal{D}_{\tau,W}$}

Initialize the parameters, $\theta_\pi, \theta_f, \phi, \psi_{\mathrm{cvae}}$

 \While{not converged}{
    Sample $(\tau^k, s_{\mathrm{g}}^k, s_{\mathrm{i}}^k)\sim \mathcal{D}_{\tau,W}$\;
    Sample $z^k \sim q_{\phi}(\cdot|\tau^k, s_{\mathrm{g}}^k, s_{\mathrm{i}}^k)$\;
    Update the model parameters by descending the batched cVAE loss $\mathcal{J}_{\mathrm{cVAE}}$ \eqref{cvae_loss}
 }
 Freeze $\theta_\pi, \theta_f, \phi$, and 
initialize the parameters $\psi_{\mathrm{flow}}$\;
 \While{not converged}{
    Sample $(\tau^k, s_{\mathrm{g}}^k, s_{\mathrm{i}}^k)\sim \mathcal{D}_{\tau,W}$\;
    Sample $z^k \sim q_{\phi}(\cdot|\tau^k,s_{\mathrm{g}}^k s_{\mathrm{i}}^k)$\;
    Compute inverse mapping of the conditional flow model, $u^k=h_{\psi_{\mathrm{flow}}}^{-1}(z^k; s_{\mathrm{i}}^k)$ and its volume distortion $v^k = \left |\det\left(\frac{\partial }{\partial z}h_{\psi_{\mathrm{flow}}}^{-1}(z^k; s_{\mathrm{i}}^k)\right)\right|$\;
    Evaluate the densities, $\log p_{\psi_{\mathrm{flow}}}(z^k|s_{\mathrm{i}}^k) $ $= \log p_{0}(u^k) +\log v^k$\;
    Update $\psi_{\mathrm{flow}}$ by descending the negative loglikelihood loss \eqref{flow_ml_loss}\;
 }
 \Output{$\theta_\pi, \theta_f, \phi,\psi_{\mathrm{flow}}$}
 \caption{Two-Stage Skill Learning}
 \label{alg:two_stage}
\end{algorithm}
\subsection{Model-Based Skill Planning}
Given the learned skill-policy, -dynamics and -prior network, we aim to construct an optimal skill-level policy $\pi^*(z|s)$ for arbitrary downstream task, c.f.,  \eqref{skill_MDP_prob}.
In this study, we intend to compare two class of model-based planning methods, online model predictive control(MPC) and offline model-based reinforcement learning(RL), each of which have the pros. and cons. depending on the usage and the characteristics of the given task and system. While any model-based planning methods are directly applicable, below we would like to make some general remarks on the practical aspects of our skill-based framework in conjunction with any online as well as offline planning methods.

\subsubsection{Online MPC}
Among other reasons, online planning methods such as MPC, i.e., decision-time computation of the optimal feedback plans, are particularly appealing over offline counterparts in that they allow direct execution of any given tasks without any further training process required.
MPC realizes suboptimal feedback plans
by repeatedly solving for a finite receding-horizon trajectory optimization problem at every time steps. Nonlinear and high-dimensional nature of the trajectory optimization problem typically limits the use of MPC for real-time applications. Model predictive path integral (MPPI) control \cite{williams2016aggressive}, one of the most popular nonlinear MPC algorithms, alleviates many of the difficulties by allowing fast, GPU-accelerated, sampling-based computation of the optimal trajectories. A direct application of the MPPI algorithm in the learned skill space is as shown in Algorithm \ref{alg:mppi}.

Still, it should be noted that the performance of online planners are fundamentally limited by its increasing online computation cost with respect to, e.g., number of sample evaluations, planning horizon length, model size. For instance, one might have to sample less or reduce the planning horizon at the cost of using more complex models, or vice versa.

\begin{algorithm}[t]
\DontPrintSemicolon
\SetAlgoLined
\SetKwInOut{Input}{Input}
\SetKwInOut{Given}{Given}
\Input{skill-policy, dynamics, prior:  $\pi_{\theta_\pi}, f_{\theta_f}, h_{\psi}$, $H=40, N=2000, u_{\mathrm{max}}=1$}
\Given{initial state $s_0$,and reward function $R(s, z)$}
Initialize input $u_t^*$, $t = 1, \cdots, H-1$

\While{not done}{
$t\leftarrow 0$ and 
$s_0^k \leftarrow s_t$ for $k=1, \cdots, N$

\For{$t=0:H-1$}{
  \ForPar{$k=0:N$}{
  Sample $N$ skills:
  $u_t^k \sim \mathrm{Uniform}(-u_{\mathrm{max}}, u_{\mathrm{max}})$
  Predict next states and evaluate the rewards:  $s_{(t+1)W}^{k}=f_{\theta_f}(s_{tW}^k,z_t^k)$,  $R_t^k = R(s_{tW}^k,z_t^k)$\;
  }
 }
 Update optimal state-skill trajectory $\{s_{tW}^*, u_t^*\}_{t} \leftarrow \mathrm{MPPI}(\{s_{tW}^k, u_t^k, R_t^k\}_{t,k})$\;
$z_0^* = h_{\psi}(u_0^*;s_0^*)$\;
\For{$t=0:W$}{
  Select action using skill-policy, $a_t = \pi_{\theta_\pi}(s_t, s_{W}, z_0^*)$\; 
  Update next state from environment\;
 }
 
 }
 \caption{Online Skill Planning: MPPI}
 \label{alg:mppi}
\end{algorithm}
\begin{algorithm}[t]
\DontPrintSemicolon
\SetAlgoLined
\SetKwInOut{Input}{Input}
\SetKwInOut{Output}{Output}
\SetKwInOut{Given}{Given}
\Input{Offline dataset $\mathcal{D}_{\tau,W}$, skill-policy, dynamics, prior:  $\pi_{\theta_\pi}, f_{\theta_f}, p_{\psi_{\mathrm{flow}}}$, $\gamma=0.96 ,u_{\mathrm{max}}=1$}
\Given{Reward function $R(s,z)$}
Initialize replay buffer $\mathcal{D} = \varnothing$, policy $\pi_{\theta_u}$\;
\While{not done}{
Sample $H$ consecutive trajectories, $(\tau^k, s_{\mathrm{g}}^k, s_{\mathrm{i}}^k)_{k=1}^{H}\sim \mathcal{D}_{\tau,W}$\;
Sample $z^k \sim q_{\phi}(\cdot|\tau^k,s_{\mathrm{g}}^k s_{\mathrm{i}}^k)$, for $k=1, \cdots, H$\;
Compute inverse mapping $u^k = h^{-1}_{\psi}(z^k;s^k_{\mathrm{i}})$ and reward $R^k = R(s_{\mathrm{i}}^k,z^k)$, for $k=1, \cdots, H$\;
$\mathcal{D} \leftarrow \mathcal{D}\cup(s_{\mathrm{i}}^k, s_{\mathrm{g}}^k, u^k, R^k)_{k=1}^{H}$\;
\If{goal-conditioned}{
$\mathcal{D}\leftarrow\mathcal{D}\cup \mathrm{HER}((s_{\mathrm{i}}^k, s_{\mathrm{g}}^k, u^k, R^k)_{k=1}^{H})$\;
}
$\bar{s}_\mathrm{i}^1 = s_{\mathrm{i}}^1$\;
\For{$t=0:N_m-1$}{
Sample base skill, $\bar{u}^t \sim \pi_{\theta_u}(\bar{s}_\mathrm{i}^t)$\;
Clamp $\bar{u}^t \leftarrow u_{\mathrm{max}}\cdot \mathrm{Tanh}(\bar{u}^t)$\;
Compute forward mapping $\bar{z}^t = h_{\psi}(\bar{u}^t;\bar{s}_\mathrm{i}^t)$\;
Predict next state $\bar{s}_\mathrm{i}^{t+1} = f_{\theta_f}(\bar{s}_\mathrm{i}^t, \bar{z}^t)$, and evaluate reward $\bar{R}^t = R(\bar{s}_{\mathrm{i}}^t,\bar{z}^t)$\;
}
$\mathcal{D} \leftarrow \mathcal{D}\cup(\bar{s}_{\mathrm{i}}^k, \bar{s}_{\mathrm{g}}^k, \bar{u}^k, \bar{R}^k)_{k=1}^{N_m}$\;
\If{goal-conditioned}{
$\mathcal{D}\leftarrow\mathcal{D}\cup \mathrm{HER}((\bar{s}_{\mathrm{i}}^k, \bar{s}_{\mathrm{g}}^k, \bar{u}^k, \bar{R}^k)_{k=1}^{N_m})$\;
}
Update $\pi_{\theta_u} \leftarrow \mathrm{SAC}(\theta_u, \mathcal{D})$\;
}
 \caption{Offline Skill Planning: Offline RL}
 \label{alg:offlineRL}
\end{algorithm}

\subsubsection{Offline RL}
Precomputation of the optimal feedback plans can be more suitable when the associated models require relatively large computational cost, or when the given task requires sufficiently sophisticated, long-horizon planning strategy that can not be easily computed in real-time. In this study, we are particularly interested in offline planning methods requiring no further repeated interaction with the environment. Offline reinforcement learning exactly tackles this problem where the objective is to extract the optimal policy from the given fixed past experience dataset. In Algorithm \ref{alg:offlineRL}, we present how the learned skill models are used with Soft Actor-Critic (SAC), state of the art off-policy RL algorithm, together with hindsight experience replay (HER) \cite{marcin2017hindsight} for tackling goal-conditioned downstream tasks. 

We note that making use of the accurate learned skill prior model largely alleviates the issue of distribution shift by explicitly rejecting out-of-distribution skill samples \cite{levine2020offline}. Specifically, we learn the feedback policy $\pi_{\theta_u}(u|s)$ that share the same support as the base skill distribution $p_0$\footnote{we use uniform base distribution, i.e., $p_0=\mathrm{Uniform}(-1,1)^{\mathrm{dim}(\mathcal{Z})}$, and hence results in compact box-type support.} and retrieve the skill $z$ via the learned forward mapping $h_{\psi}$.
Other offline RL algorithms can also be applied in both model-based and model-free settings in a similar fashion. Model-based methods are likely to work better than the model-free ones under limited dataset size and sufficiently accurate model \cite{yu2021combo}.

\begin{figure*}[t]
\centering
\includegraphics[width=1.8\columnwidth]{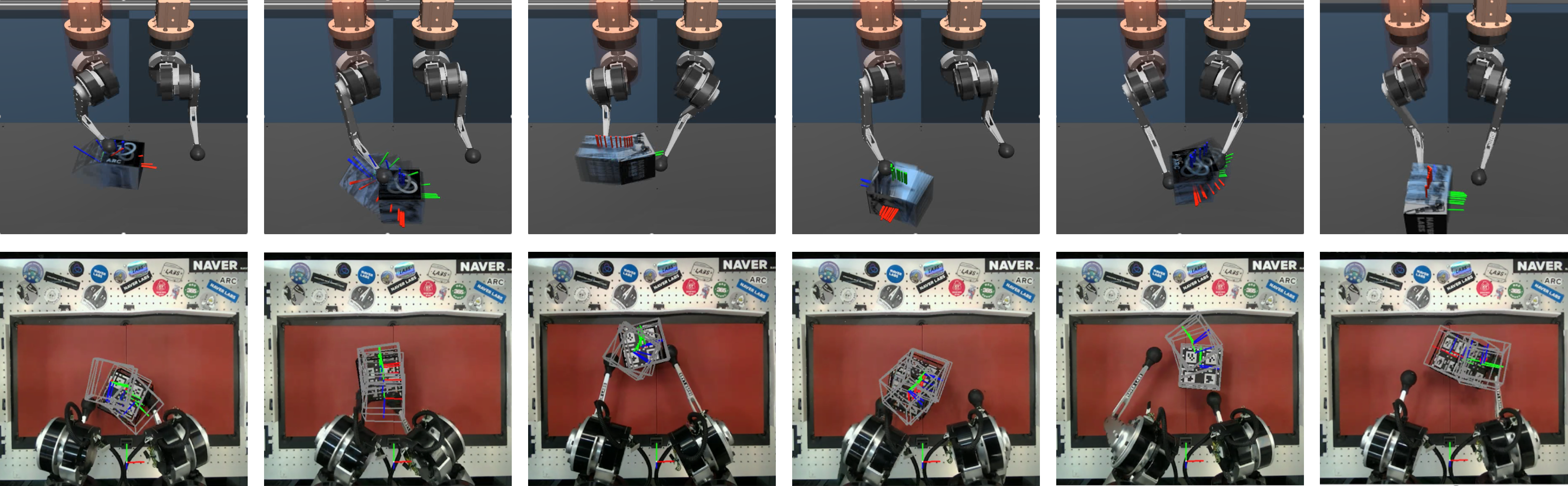}
\caption{Random skill sequence execution, and future trajectory prediction for duration of $H=10$ steps (1 second). Diverse and predictable manipulation behaviors such as flipping, dual-arm pushing, single-arm pushing, rotating are generated from the model.}
\label{fig:mini_random_skill_snapshot}
\end{figure*}
\section{EXPERIMENTS}
In this Section, we present extensive simulated and real experimental results to assess the efficacy of the proposed skill learning framework and also to validate the claims that were raised throughout the paper. 

\begin{figure}[t]
    \centerline{
    \subfigure[Demonstration]
    {
    \includegraphics[width=0.35\columnwidth]{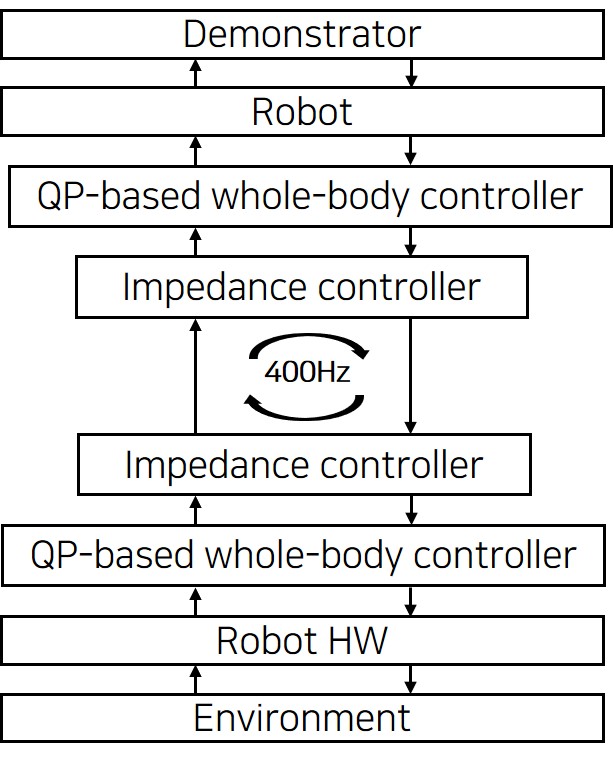}%
    \label{fig:bilateral_archi}}
    \hfil ~~
    \subfigure[Test]
    {
    \includegraphics[width=0.35\columnwidth]{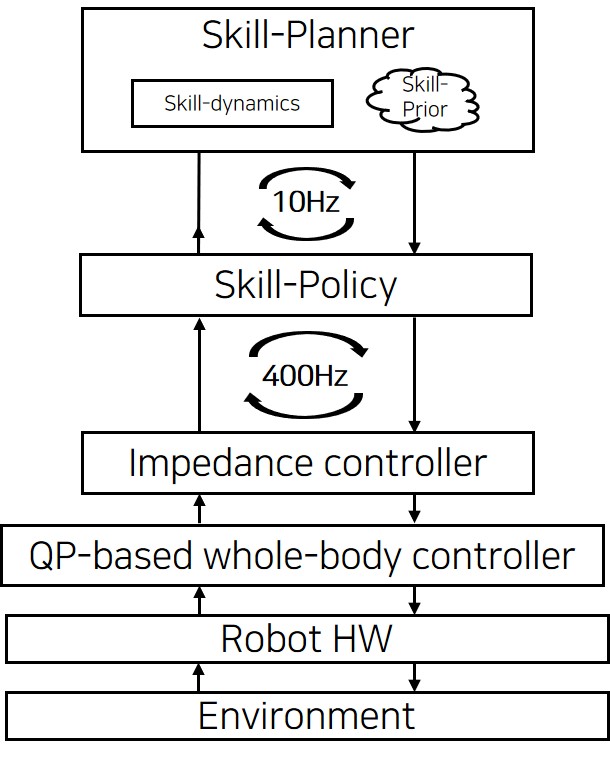}%
    \label{fig:skill_archi}}}
\caption{Control system architectures. For MPC planning, the skill planner and skill policy loop is run asynchronously in a parallel manner due to the large computational burden of the MPC planner. Without such implementation, i.e., using naive serial computation, the time delay incurred by the MPC planner module resulted in the system instability in both real-time simulation and real-world.}
\label{fig:control_architecture}
\end{figure}

As shown in Figure \ref{skill_hierarhcy}, we verify our algorithm on a torque-controllable dualarm robot (4-dof for each arm actuated by the commercial proprioceptive actuators, T-motor AK60-9), with RGBD camera, realsense L515, mounted from the top-view. 
For demonstration dataset collection, the human demonstrator gets to physically operate the end-effectors of same pair of robot arm bilaterally coupled with either simulated or real slave robot, and is also provided with visual feedback from the camera. The overall control system architecture is as shown in Figure \ref{fig:control_architecture}. The demonstrator is asked to continuously produce random, yet, diverse and dexterous box manipulation behaviors, e.g., flipping, rotating, pushing either with both arms or a single arm. The box used in both simulated and real experiment weighs 1kg with dimension of 0.1m $\times$ 0.1m $\times$ 0.1m. Four sets of Apriltag3 markers are attached at each side of the box for tracking its position and orientation feedback information in real-time (at 30Hz). In both simulated and real demonstrations, the total duration of the gathered demonstration data is about 2 hours in physical time, which has been collected over 4 separate trials (equally for about 30 minutes each).

The downstream task objective at test time is to configure the box to a randomly given desired position and orientation. The task success condition is set to require the box position error to be less than $1.5 \mathrm{cm}$ and the orientation error to be less than $20 \mathrm{deg}$.
We model skill-policy network as Recurrent Neural Network (RNN), skill-posterior as bidirectional-RNN, and skill-dynamics network as Multilayer Perceptron (MLP). For cVAE, the mean and variance for conditional Gaussian skill-prior model are modeled as MLP. For flow-based skill-prior modeling, we use Inverse Autoregessive Flow (IAF) network (of 2-block and 2-layer) for enabling fast forward sampling \cite{kingma2016improved}. Throughout all the experiments, the minimum unit duration of a skill is set as $W=40$ which corresponds to time duration of 0.1 second. As noted in Remark \ref{rem:skill_extension}, Figure \ref{fig:mini_random_skill_snapshot} depicts randomly sampled skill sequences for 10 steps (duration of $1$ second), showing diverse and ``predictable'' manipulation behaviors.
For every MPC trials, we set the planning horizon as $4$ seconds ($H=40$), number of trajectory samples as $N=2000$, and the reward function is simply given as scaled Euclidean distance to the target box position and orientation. For offline RL, we use a sparse goal-conditioned reward, i.e., $R=0$ if success, and otherwise $R=-1$. Maximum episode length is set to 50 seconds.

\subsection{Simulation} \label{sec:exp_sim}
\begin{figure}[t]
\centering
    \subfigure[Performance over models]
    {
    \includegraphics[height=4.7cm]{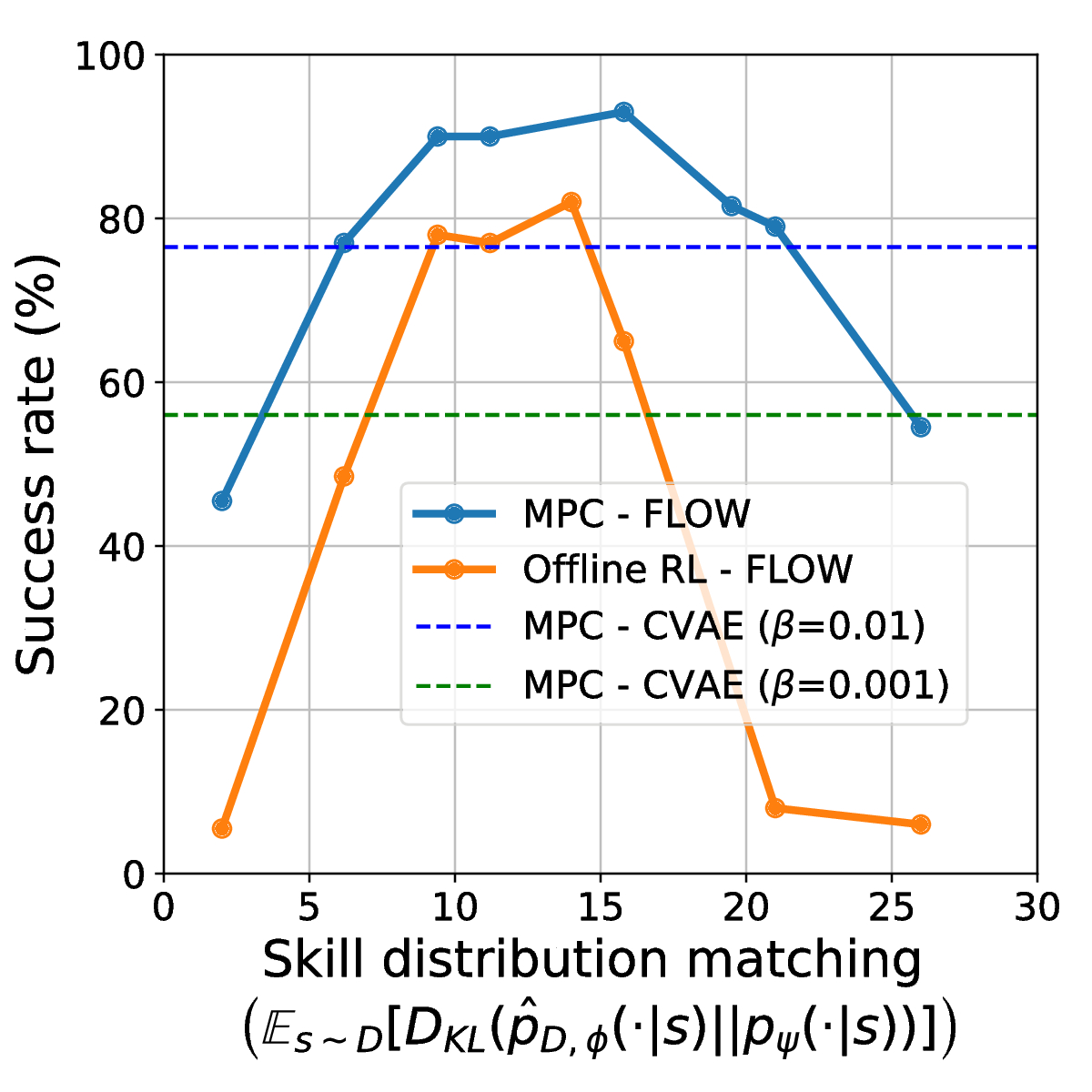}%
    \label{fig:distribution_vs_performance}}
    \hfil
    \subfigure[Performance over planners]
    {
    \includegraphics[height=4.7cm]{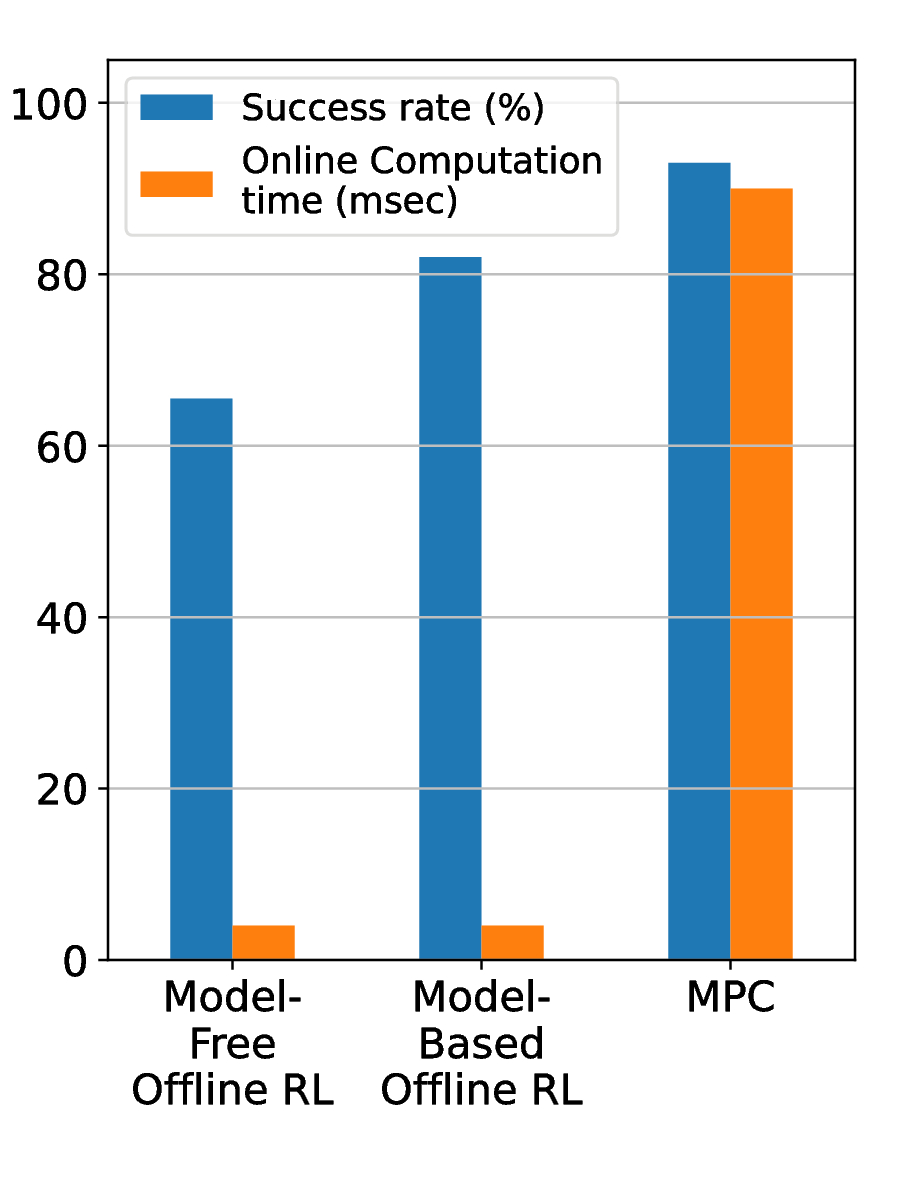}%
    \label{fig:planner_vs_performance}}
\caption{Simulation experiment results averaged over 200 random-goal episodes.
The metric $\mathbb{E}_{\mathcal{D}}[D_{KL}]$ is estimated by offsetting the likehood loss by the estimated conditional entropy of the data distribution. The conditional entropy is obtained from the converged likehood loss using the sufficiently high-capacity (16-layer) flow model.}
\label{fig:simulation_performance}
\end{figure}

In the simulation experiment, we first aim to identify the impact of generative skill modeling performance to the downstream skill planning performance. More specifically, we show how the proximity of the learned skill distribution, $p_{\psi}(\cdot|s)$, to the offline data distribution, $\hat{p}_{\mathcal{D},\phi}(\cdot|s)$ can influence the task success rates of the model-based planners.
Figure \ref{fig:distribution_vs_performance} shows the plot of task success rates of model-based planners, MPC and Offline RL, over using different choices of skill-prior models differing in the amount distribution match. All the skill-prior models used, except for the one having the lowest value of $\mathbb{E}_{\mathcal{D}}[D_{KL}(\hat{p}_{\mathcal{D},\phi}\|\|p_{\psi})]$, have the identical architecture with different training epoches. The model with the lowest $\mathbb{E}_{\mathcal{D}}[D_{KL}]$ is obtained by converging sufficiently high-capacity model (16-block 2-layer IAF); note however that this model could not be used for real-time application of MPC planner due to the large online computational burden, and hence only for this model the planning performance is evaluated in the non-real-time simulation environment. The rest of the performance metrics are all measured in the real-time simulation environment. It can be clearly observed that the planning performance is largely dependant on the quality of distribution match on the learned skill representations. There exists a sweet-spot, $\mathbb{E}_{\mathcal{D}}[D_{KL}] \sim 10-15$, at which the learned distribution is neither over-fitted nor under-fitted to the data distribution. Not surprisingly, under-fitted models (large $\mathbb{E}_{\mathcal{D}}[D_{KL}] > 20 $) show low task success rates by producing many ``mistakes''; more precisely, they tend to produce many out-of-distribution samples at which the predictive models, i.e., skill-policy and skill-dynamics, fail to generalize well. Perhaps, more interestingly, when the model over-fits the data distribution ($\mathbb{E}_{\mathcal{D}}[D_{KL}] \sim 1 - 5$), the performance start to degrade more sharply. We could observe that the robot behaves very conservatively in these cases, frequently producing far-from-optimal behaviors to configure the box. We would argue that over-fitted models try to pose very low density to even nearby skills to the data points at which the predictive models are supposed to fairly generalize well.

Now, let us compare the results between two different types of model-based planners, online MPC and offline RL. Before discussing about the clear performance gap between MPC and offline RL, we would like to first touch on the fact that (in the present experiment) the performance of offline RL is a lot more sensitive to the variations in the distribution match than that of MPC. It is fairly well-understood that offline RL can be very susceptible to the distribution-shift. Generally, one should care about distribution-shift problem that arise in both state and action spaces. Majority of the offline RL algorithms are designed to mitigate distribution shift on action spaces \cite{levine2020offline}. Policy-constraint methods in which the present method also lies, explicitly constrains the target policy to the learned behavior policy. Another body of recent approach referred to as conservative Q-learning more directly controls the action distribution-shift problem by appropriately regularizing the Q function. For the present experiment, the method showed minor performance improvement. 
That being said, we would argue that the present performance gap between the MPC and offline RL comes in large part to the quality and diversity of state-space coverage \cite{keller2020model} on which the model is being evaluated or trained. It is noted that online planning methods like MPC gets the chance to directly try out predictive computation on the true states being observed at test time. On the other hand, it is not as obviously the case for offline learned models to be trained on the states that are not contained in the fixed offline dataset, unless one can generate additional synthetic rollout samples using learned dynamic models. 
Indeed, Figure \ref{fig:planner_vs_performance} shows that using model-based offline RL improves the performance to some extent as compared to model-free setting where only the offline dataset is used to train the policy. Although it is out of scope of this paper, we anticipate that using MPC plans to continually collect a more diverse, real, and safe state-covering data can be a promising direction in which to boost the performance and practical usability of data-driven offline RL methods.
\begin{figure}[t]
\centering
\includegraphics[height=5.2cm]{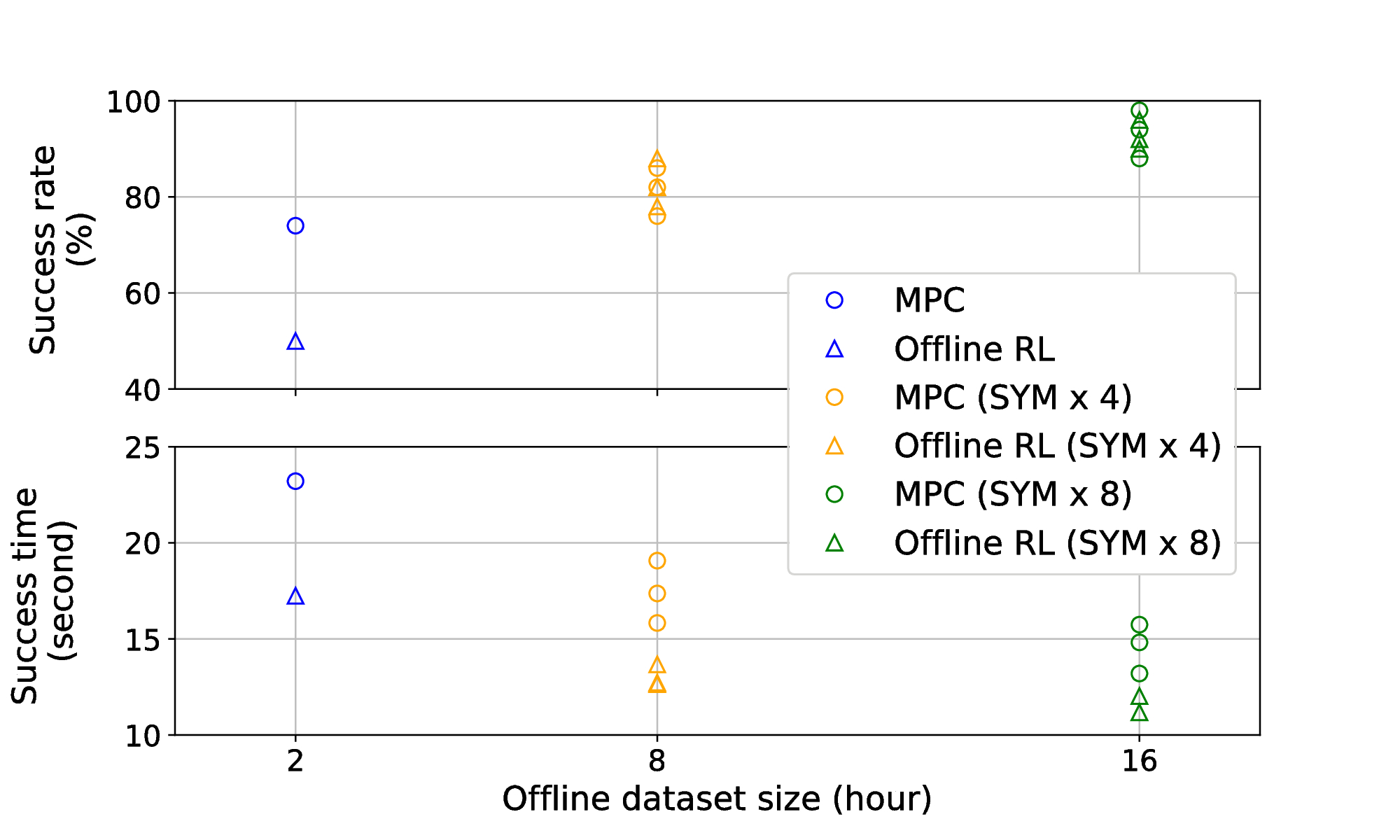}
\caption{Real-world experiment results averaged over 50 random episodes.}
\label{fig:realworld}
\end{figure}
\subsection{Real Hardware}
In the real hardware experiment, we have mostly followed the skill modeling and experiment design decision/rules made in the previous simulation studies. Figure \ref{fig:realworld} shows the experimental results obtained from the skill models an planners that achieve the best downstream task success rates, while also increasing the size of offline demonstration dataset. We note that we actually scale the dataset size by augmenting the 2 hours of play data by exploiting the object symmetry, not by physically collecting more demonstrations\footnote{the cube-shape box used in the experiment is symmetric under at most 24 set of rotations.}. First of all, it can be observed that under the same amount of offline dataset as in the simulation study, i.e., 2hours of play demonstrations, the overall task success rate, with both MPC and offline RL, is substantially reduced in the real-world setting. Nonetheless, increasing the scale of offline dataset clearly improves the overall downstream skill planning performance. We also note that as opposed to the previous simulation study, the performance of offline RL showed little difference between using model-based and model-free SAC, given 2 hours of offline dataset. Perhaps, more interestingly, as the dataset size scales over 8 hours, model-free SAC actually outperformed model-based one. We would argue that as real data becomes sufficiently diverse and dense, error-prone synthetic model rollout data could have higher chance of making conflict with real-world data. In this regard, we expect explicit consideration of conservatism toward synthetic dataset could be an effective remedy to the problem \cite{yu2021combo}.
\section{CONCLUSION}
We have shown that the idea of offline learning of model-based skills from haptic, teleoperated play demonstrations can be an efficient, practical means to teach a robot perform complex, real-world contact-rich manipulation tasks. The proposed method is validated through extensive simulated and real-world experiments, while also verifying key compounding factors, e.g., distribution match, dataset scale, types of planning algorithms, that pertain to the downstream task performance.

We believe there can be many ways in which to further improve the performance and generality of the offline skill learning approach. One core direction remains answering how should {\em skill} be defined. For instance, if one wishes to build a robot that can paint on a canvas, a unit skill may be more efficiently encoded as a single stroke motion subject to variable duration. Also, skills can be defined to be associated with not just initial and final state, but more generally with starting and goal set conditions, as is widely studied in the literature, e.g., option framework or feedback-motion-planning. Regardless, the usefulness of any type of skill should be judged not only by its algorithmic effectiveness, but also by feasibility of large-scale data-driven learning; that is, the required training data should be easily collected and labeled at scale in practice.

It becomes more evident that the value of diverse and rich real-world data is hard to overemphasize in robot learning. The use of haptic teleoperation interface and self-supervision techniques adopted in the present study can be the part of effort toward reducing the data collection cost in the real-world. Further reducing the data collection cost would be to automate the exploration strategy \cite{sharma2019dynamics, openai2021asymmetric}.  In this regard, as discussed in Section \ref{sec:exp_sim}, using direct MPC plans to realize exploration strategy appears to be one appealing solution to mitigate the state-coverage problem in offline RL and also to robustify the skill model.

Lastly, we believe that there also remains great potential in efficient usage of limited data by exploiting the intrinsic structure therein. The case of simple symmetry-driven data augmentation strategy demonstrated in the real-world experiment shows the potential impact of exploiting the geometric structure in data. Symmetric properties are prevalent in manipulation as it happens in physical world. Investigation of skill models admitting invariance properties under certain types of group transformations, e.g., geometric deep learning models that operate on shapes, point clouds, etc., appears to be a profitable direction of future research.



\bibliographystyle{IEEEtran}
\bibliography{IEEEabrv,reference}

\end{document}